\documentclass[conference]{IEEEtran}
\IEEEoverridecommandlockouts
\usepackage[utf8]{inputenc}
\usepackage{color}
\usepackage{pbox}
\usepackage{booktabs}
\usepackage{amsmath,amssymb}
\usepackage{amsthm}
\usepackage{verbatim} % for comments
\usepackage{bm}
\usepackage{graphicx}

\newcommand{\squishlist}{
   \begin{list}{$\bullet$}
    { \setlength{\itemsep}{2pt}    \setlength{\parsep}{0pt}
      \setlength{\topsep}{5pt}     \setlength{\partopsep}{0pt}
      \setlength{\leftmargin}{1.35em} \setlength{\labelwidth}{1em}
      \setlength{\labelsep}{0.5em} } }

\newcommand{\squishend}{
    \end{list}  }

\usepackage{hyperref}
\usepackage{mathtools}

\newcount\Comments  % 0 suppresses notes to selves in text
\usepackage{helvet}

\usepackage{subcaption}
\usepackage{longtable} % for the end of the paper

\begin{document}

\title{Detection and Mitigation of Rare Subclasses in Deep Neural Network Classifiers\\
}

\author{\IEEEauthorblockN{Colin Paterson}
\IEEEauthorblockA{\textit{University of York, York, UK}\\
colin.paterson@york.ac.uk}
\and
\IEEEauthorblockN{ Radu Calinescu}
\IEEEauthorblockA{\textit{University of York, York, UK} \\
radu.calinescu@york.ac.uk}
\and
\IEEEauthorblockN{Chiara Picardi}
\IEEEauthorblockA{\textit{University of York, York, UK} \\
chiara.picardi@york.ac.uk}
}

\maketitle

\begin{abstract}
Regions of high-dimensional input spaces that are underrepresented in training datasets reduce machine-learnt classifier performance, and may lead to corner cases and unwanted bias for classifiers used in decision making systems. When these regions belong to otherwise well-represented classes, their presence and negative impact are very hard to identify. We propose an approach for the detection and mitigation of such \emph{rare subclasses} in deep neural network classifiers. The new approach is underpinned by an easy-to-compute \emph{commonality metric} that supports the detection of rare subclasses, and comprises methods for reducing the impact of these subclasses during both model training and model exploitation. We demonstrate our approach using two well-known datasets, MNIST's handwritten digits and Kaggle's cats/dogs,  identifying rare subclasses and producing models which compensate for subclass rarity. In addition we demonstrate how our run-time approach increases the ability of users to identify samples likely to be misclassified at run-time.
\end{abstract}

\section{Introduction}

Class imbalance is a major and widely recognised problem in machine learning (ML)  \cite{guo2008class,Japkowicz00theclass}, and significant research effort has been dedicated to its mitigation, e.g.\ \cite{batuwita2010fsvm,buda2018systematic,chawla2004special,lemaitre2017imbalanced,liu2008exploratory}. In contrast, its surreptitious cousin \emph{subclass rarity} is far less understood and addressed by existing research. A \emph{rare subclass} is an underrepresented region of a class where this class is otherwise well represented in the datasets used to train and test an ML classifier. Rare subclasses are known to lead to unwanted bias~\cite{d2017conscientious} and other corner cases~\cite{huang2018challenges} when present in datasets used to train decision-making classifiers. 
As an example, in applications such as medical diagnosis, the progression of a disease does not always follow the same pattern~\cite{olivotto2012patterns}. Rare forms of disease progression may possess features which are less common in the training and testing sets, leading to undue confidence in diagnosis decisions when ML is employed with respect to these rare forms of the disease.

Methods have been proposed to address this problem for the scenario where the features of the rare subclass can be anticipated because they correspond to gender, age, race, religion or other \emph{protected attribute(s)}~\cite{bellamy2018ai} that might induce discrimination, e.g., \cite{calmon2017optimized,7961993}. However, these methods are ineffective for rare subclasses not associated (or not directly associated) with protected attributes. Many potential examples of such rare subclasses exist, e.g.\ hand-written digits when samples from users suffering from Parkinson's disease are not recorded~\cite{thomas2017handwriting}, or medical data which does not include samples from women undergoing menopause~\cite{westergaard2019population}.

Whilst we may, and should, examine misclassified images at training time to identify systematic failures of the learnt model, this is not usually possible when the ML classifier is used for decision making on unlabelled data. Furthermore, for many applications such an examination would be either too labour intensive or, where undertaken, may  reveal little about the features used to classify samples.

Our paper proposes an approach for the detection and mitigation of these types of rare subclasses in deep neural network (DNN) classifiers. The new approach comprises:

\vspace*{-1mm}
\squishlist
\item[1)] A method for the efficient computation of a  \emph{commonality metric}. Applied to a sample from the test set or an unlabelled sample being classified by the DNN, this metric indicates how frequently data samples with similar characteristics were encountered in the training dataset.
\item[2)] A method that applies our new metric to the test dataset, to detect rare subclasses, supporting the augmentation of the training dataset with additional samples representative of these subclasses. Used during the model learning stage of ML lifecycle~\cite{ashmore2019assuring}, this method improves the DNN performance both for the (previously) rare subclass and overall.
\item[3)] An online method that applies our commonality metric to unlabelled samples being classified, to identify samples with characteristics potentially unseen during training. Obtaining a second opinion for the few such samples (from an alternative, higher-cost classifier such as a human operator) can significantly reduce the number of classification errors.
\squishend

\vspace*{-1mm}
The rest of the paper is organised as follows. Section~\ref{sec:background} provides the necessary background on DNNs. Section~\ref{sec:Approach} presents our approach by  describing the calculation of the commonality metric, and  the use of this new metric to mitigate against the effects of rare subclasses during training and at run-time. Section~\ref{sec:Evaluation} evaluates our approach using the MNIST handwritten digits data set and a Kaggle competition data set for a binary classifier applied to images of cats and dogs.
We start by showing that our commonality metric is correlated with misclassification rates. Next we demonstrate how our approach supports the creation of DNN classifiers which mitigate the effects of subclass rarity. Finally, we show how our approach allows for improved identification of misclassified samples at run-time.
We consider the body of research within which this work fits in Section~\ref{sec:Related}, before concluding and suggesting directions for future work in Section~\ref{sec:Conclusions}.

\section{Background \label{sec:background}}

A deep neural network (DNN) is a parameterised function $y = f(\mathbf{x}; \mathbf{\theta})$ that maps a vector  $\mathbf{x} = (x_1, x_2,  \ldots, x_m)$ of $m$ input features to an output $y$, where $\mathbf{\theta}$ is a set of parameters. During training, a set of $N$ labelled input samples $\{(\mathbf{x}_i,y_i)\}_{1\leq i\leq N}$ is available, and the DNN parameters $\theta$ are learnt such that a loss function that measures the prediction errors (i.e., the differences between $f(\mathbf{x}_i; \mathbf{\theta})$ and $y_i$ for all $1\!\leq\! i\!\leq\! N$) is minimised.

A DNN is structured as a feedforward network comprising $L$  layers of computation with the output of each layer forming the input to the next such that:
\begin{equation}
    y = f^{(L)}\left( \cdots f^{(2)}\left( f^{(1)}  (\mathbf{x})  \right) \cdots \right)
    \label{eq:DNN}
\end{equation}
where the integer in the suffix denotes the layer.

For modern classification tasks, layers with different computation structures are combined in~\eqref{eq:DNN}. These include convolutional layers able to learn important image features, max pooling layers that reduce the dimensionality of the problem space, and fully connected layers where all outputs from the previous layer are combined and a non-linear activation function is applied to the result. The overall aim is to extract the most important information from the input such that the output from the penultimate layer may be combined to produce a prediction of the most likely class. For multiclass classification problems, this prediction is typically provided by a softmax activation function~\cite{chollet2018deep} which normalises the output to provide a probability distribution across $k\geq 2$ output classes.

\section{Approach\label{sec:Approach}}

\begin{figure*}
	\centering
	\includegraphics[width=0.85\textwidth]{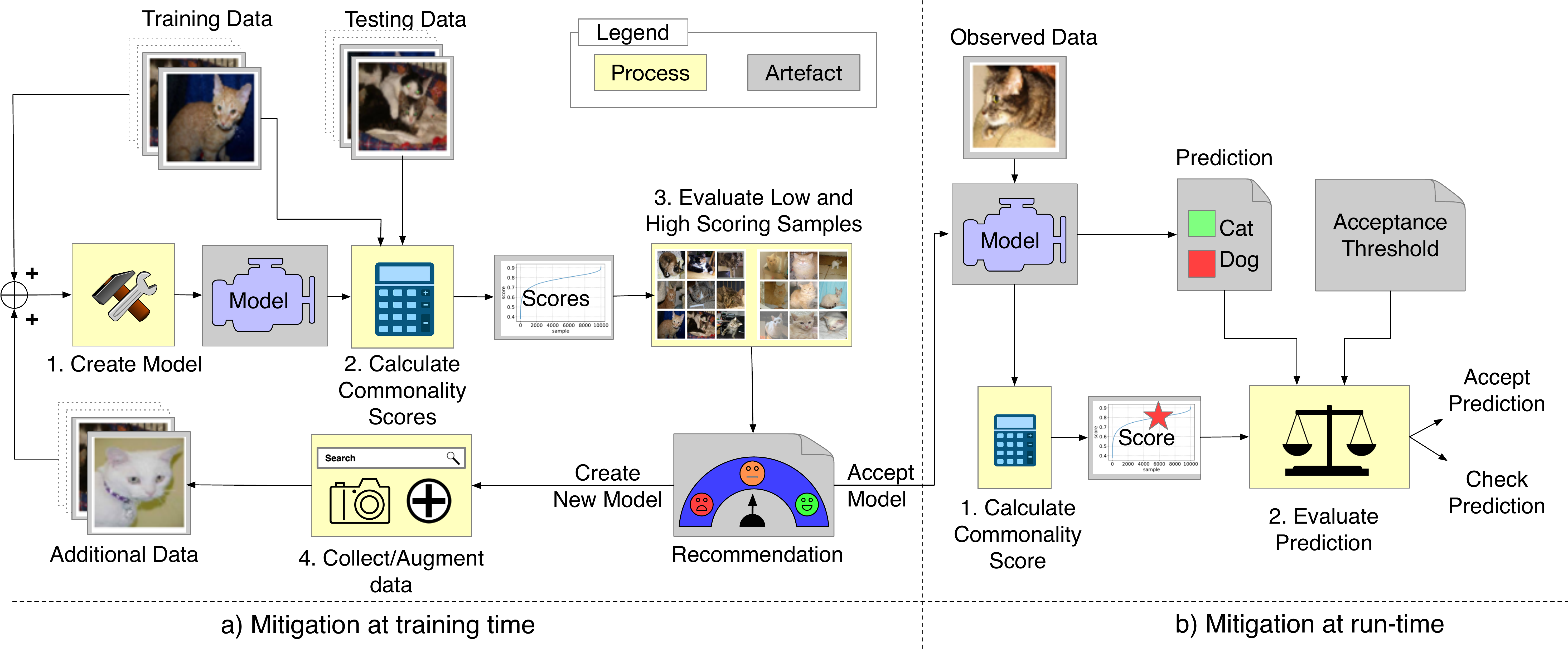}
	
	\caption{Processes for mitigating rare subclasses at training time (left) and run-time (right).
		\label{fig:Process}}
		\vspace{-3mm}
\end{figure*}

In this section we describe the methods underpinning our approach for detecting and mitigating the effects of rare subclasses during DNN classifier development and at run-time. We start by introducing the \emph{commonality score} metric that we use to identify rare subclasses. We then describe how this score can be used, firstly at training time and then at run-time, to carry out mitigation activities that reduce the impact of rare subclasses on system performance.

\subsection{Commonality Score Computation \label{ssec:score}}

Our commonality score considers the activation of the neurons from the penultimate layer of a DNN classifier. Importantly, it can be calculated both for a labelled data sample from the testing dataset (during the testing of the DNN), and for an unlabelled data sample (when the DNN performs online classification). In both cases, the score reflects the similarity between (i)~the activations of the penultimate-layer neurons for the sample under analysis and (ii)~the activitations of these neurons for the samples from the DNN training dataset. The intuition is that the two activations are likely to be similar for samples resembling those from the training dataset, and significantly different for samples from rare subclasses.

Our method for commonality score computation comprises two stages. 
~Given a $k$-class DNN classifier with $n$ neurons in its penultimate layer, the first stage computes a $n\times k$ \emph{cumulative activation matrix} $\mathbf{C}$ based on the training dataset $X$ used to learn the DNN. The element $c_{ij}$ from this matrix counts how many times the $i$-th neuron from the penultimate DNN layer is activated across all training samples from class $j$, i.e.
\begin{equation}
 c_{ij} = \sum_{x\in X, \mathit{label}(x)=j} \alpha_i(x),
 \label{eq:matrix}
\end{equation}
where $\mathit{label}(x)\in\{1, 2, \ldots, k\}$ represents the class of training sample $x\in X$, and $\alpha_i(x)\in\{0,1\}$ has value $1$ if the $i$-th neuron from the penultimate DNN layer is activated for sample $x$ and value $0$ otherwise. 
The cumulative activation matrix $\mathbf{C}$ only needs to be computed once for a trained DNN classifier.

The second stage of our method uses the cumulative activation matrix $\mathbf{C}$ to calculate the commonality score for a data sample $x'$, where $x'$ can be a labelled sample from the test dataset or an unlabelled sample to be classified by the DNN. Assuming that the DNN classifies $x'$ as belonging to class $j$, the commonality score is calculated using the $j$-th column of $\mathbf{C}$ (i.e.\ the \emph{cumulative activation vector} for class $j$):
\begin{equation}
    \mathit{score}(x') = \frac{\sum_{i=1}^n \alpha_i(x')c_{ij}}{\sum_{i=1}^n c_{ij}},
    \label{eq:sample-score}
\end{equation}
where (as before) $\alpha_i(x')\in\{0,1\}$ has value $1$ if the $i$-th neuron from the penultimate DNN layer is activated for sample $x'$ and value $0$ otherwise. Thus, the commonality score for $x'$ is computed as the ratio between the sum of neuron activations observed in training data for the neurons that also fired for $x'$ and the sum of all neuron activations in training data. A low score indicates an uncommon sample while a high score indicates a common sample. Importantly, $score(x')$ can be calculated efficiently in $O(n)$ time.

\subsection{Training-time Mitigation Method}

Figure~\ref{fig:Process}a illustrates our four-stage process for the mitigation of rare subclasses during training.
This process combines traditional ML training with the use of the commonality metric from Section~\ref{ssec:score} to drive the augmentation of training data such that the misclassification of rare subclasses is reduced.

In stage~1, the approach uses training data collected using traditional strategies to create a DNN model. Once a model has been constructed with acceptable performance, 
the training data are used in stage~2 to calculate the cumulative activation matrix~\eqref{eq:matrix}, which is then used to compute the commonality score~\eqref{eq:sample-score} for each sample in the testing set.

In stage 3, the testing set is partitioned into two subsets. The first subset contains those samples which have the lowest commonality scores in the set and the second contains those with the highest scoring. These two subsets are then passed to an evaluation activity whose purpose is to  identify the features that characterise a majority of images in the low-scoring sample subset. This feature identification is typically undertaken by a human worker or team of workers, by comparing images in, and between, the two subsets. The size of the subsets used is likely to be informed by the capacity of the evaluation team and the complexity of analysis of the samples. The features identified in this stage are summarised into a recommendation. 
This recommendation indicates whether no further action is required (and hence the model should be accepted) or  the training data set should be augmented. 

Finally, if the augmentation of the training data set is required, Stage 4 performs this augmentation. This can be achieved by collecting additional samples which possess the features that characterise low-scoring samples. Alternatively, the existing data set may be used to generate new samples through augmentation processes such as scaling, rotation and colour shifts, with the aim of extending the training data set with (synthetic) samples that exhibit the features of the rare subclasses.
The newly gathered and/or generated samples are then added to the training set. A new model is then be created, and the process is repeated until no actionable features can be identified for those samples which have low scores.

\subsection{Run-time Mitigation Method}

Figure~\ref{fig:Process}b shows our two-stage run-time mitigation process. This process is applicable to unlabelled samples being classified after the DNN model has been accepted for deployment. 

Given an unlabelled sample classified by the model, stage~1 of our process calculates its commonality score~\eqref{eq:sample-score} using the cumulative activation matrix~\eqref{eq:matrix} derived at training time. 
In stage~2, we then evaluate the model prediction using an acceptance threshold pre-calculated  based on the commonality scores of the  training data samples. When the score of the unlabelled sample is below this threshold, the DNN prediction is less trustworthy, and a secondary mechanism may be employed to check its validity.

We calculate the threshold $\tau$ for evaluating the trustworthiness of the DNN predictions using \emph{Tukey's fences}~\cite{tukey:1977:EDA}, an outlier detection technique that utilises the interquartile range of the commonality scores observed in the training data:
\begin{equation}
\label{eq:outlier}
\tau = Q_1 - k(Q_3-Q_1)
\end{equation}
where $Q_1$ and $Q_3$ are the upper and lower quartiles, respectively, and $k$, a tunable parameter, is typically set as 1.5.

A sample with a score below this threshold indicates that the sample is uncommon with respect to the training data used and, as such, is more likely to be misclassified. 
Therefore, a simple mitigation process is to accept any prediction with a score above the threshold, and to present samples with scores below the threshold to a second, possibly human, system to confirm the prediction or identify a misclassification.

\section{Evaluation\label{sec:Evaluation}}

This section shows the ability of our approach to identify and mitigate rare subclasses in the MNIST handwritten digits and the Kaggle cats and dogs data sets.\footnote{The two data sets and model structures we used are available from Keras at \url{https://github.com/keras-team/keras/blob/master/examples/mnist_cnn.py}, and from Kaggle at \url{https://www.kaggle.com/c/dogs-vs-cats/data}, respectively.} 
To enable the reproducibility of our results, we provide further details about the data sets, models and experiments from this section on our supporting website~\url{https://www.cs.york.ac.uk/tasp/rarity/}. 

\subsection{Experimental Setup}

In order to evaluate the ability of our approach to detect and mitigate rare subclasses we considered two different image classification data sets, and an associated model for each set. The MNIST data set comprises handwritten digits encoded as fixed size grayscale images, while the Kaggle cats and dogs data set contains full colour images of irregular size. 

Although the original Cats and Dogs image set has a test set of 12,500 images, these are unlabelled. In order to evaluate misclassification rates, we chose 5000 images at random to be hand labelled. From these, 11 images were found to be neither dogs nor cats, and were removed. This resulted in a labelled test set of 4989 images, which we used in our experiments.

We trained classifiers using the training data and used our approach to record the neuron activation frequency and calculate the cumulative activation matrix as described in Section~\ref{ssec:score}.
We then presented each of the images in the test data set to the model and calculated a commonality score for each image; the results are shown in Figure~\ref{fig:Scores}.

Figure~\ref{fig:ScoresA} shows the commonality score for all samples in the test set sorted into ascending order, while Figure~\ref{fig:ScoresB} shows this information as a box plot. We note that for both data sets a number of samples are identified as outliers, i.e. they have a significantly lower commonality score than the other samples.

\subsection{Commonality Score -- Misclassification Rate Correlation}

\begin{figure*}[t]
	\begin{subfigure}[b]{0.30\linewidth}
		\centering
		\includegraphics[width=\linewidth]{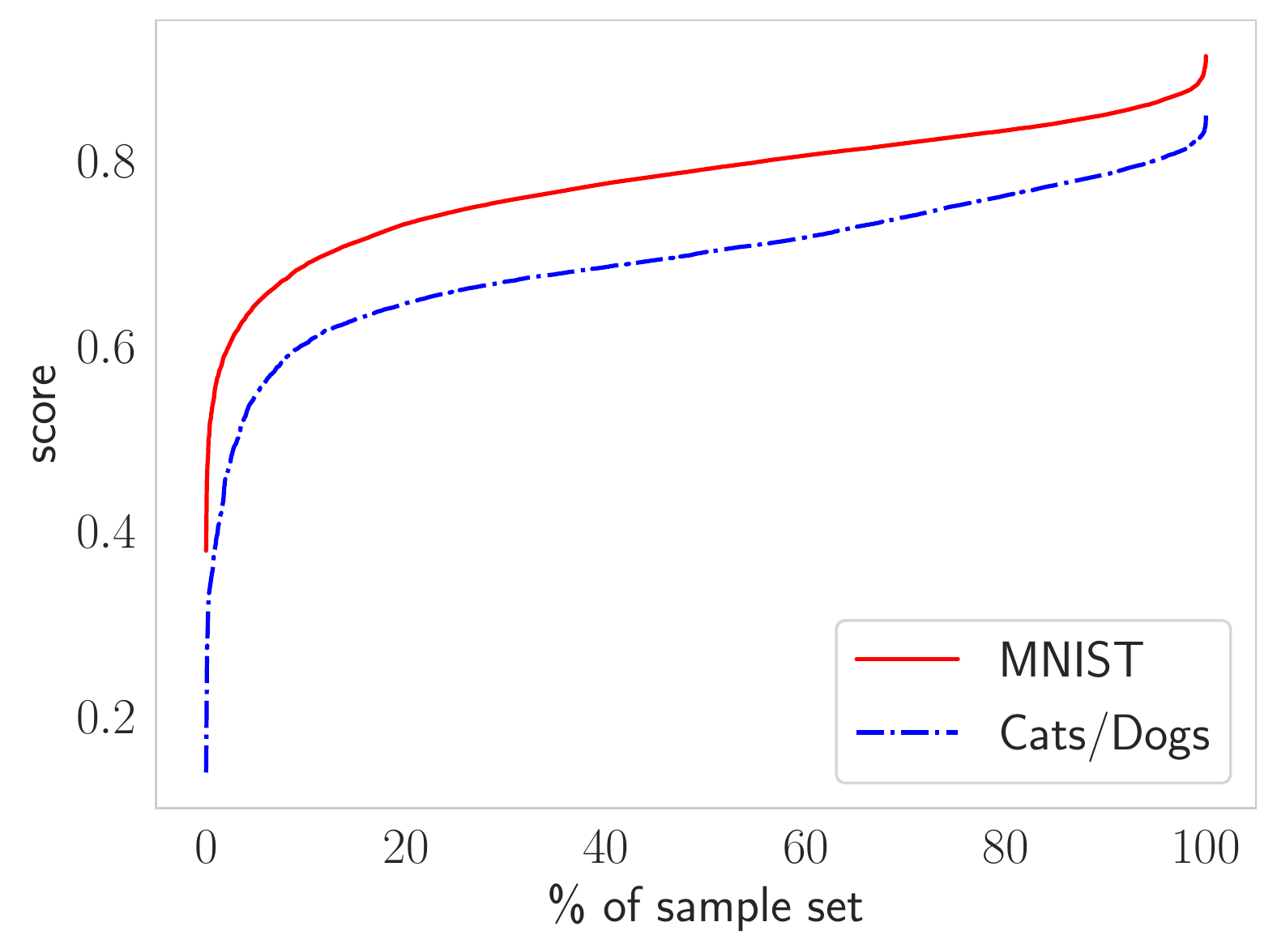}
		\caption{Score of samples in ascending order\label{fig:ScoresA}}
	\end{subfigure}
	\begin{subfigure}[b]{0.30\linewidth}
		\centering
		\includegraphics[width=\linewidth]{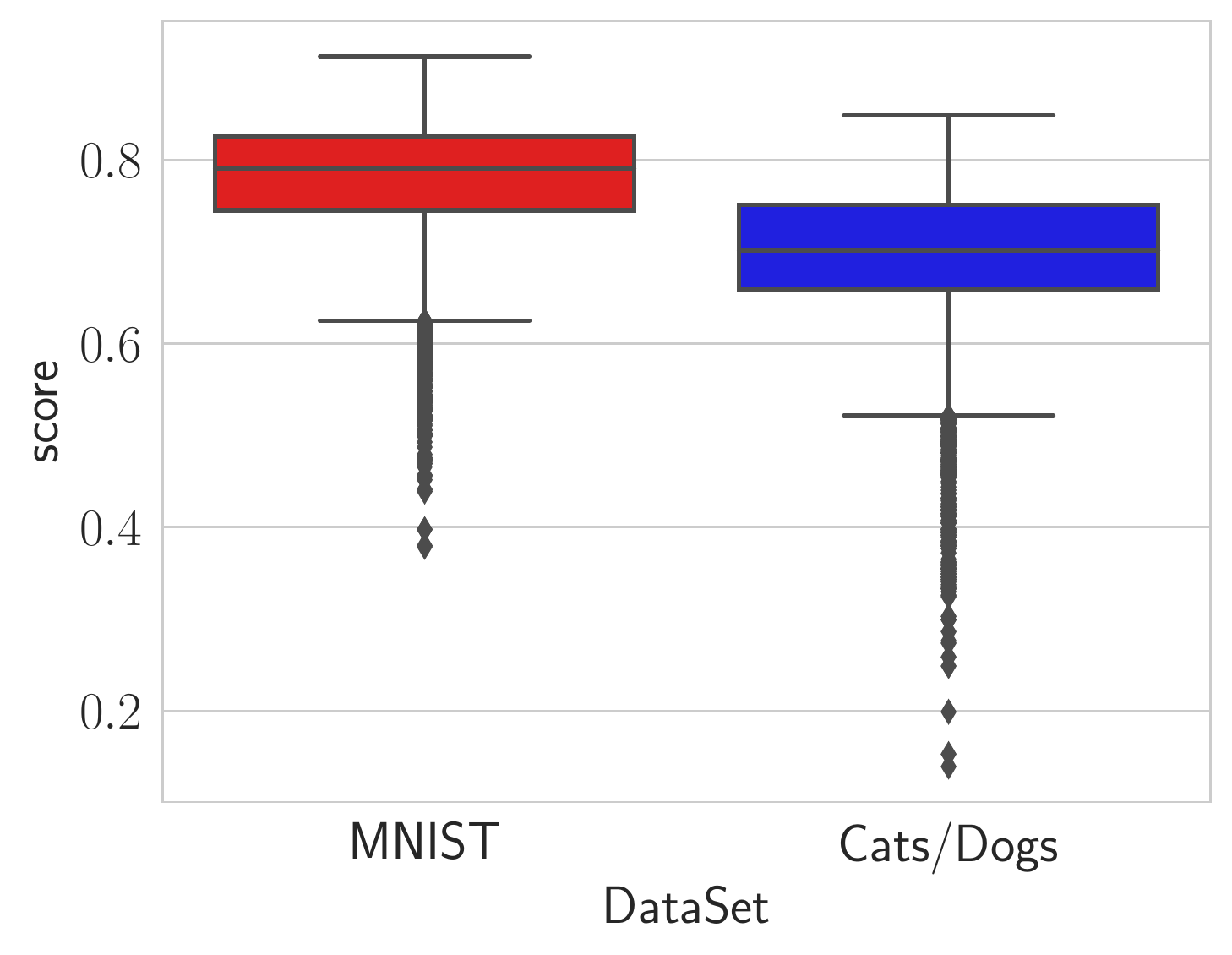}
		\caption{Commonality score with outliers\label{fig:ScoresB}}
	\end{subfigure}
	\begin{subfigure}[b]{0.30\linewidth}
		\centering
		\includegraphics[width=\linewidth]{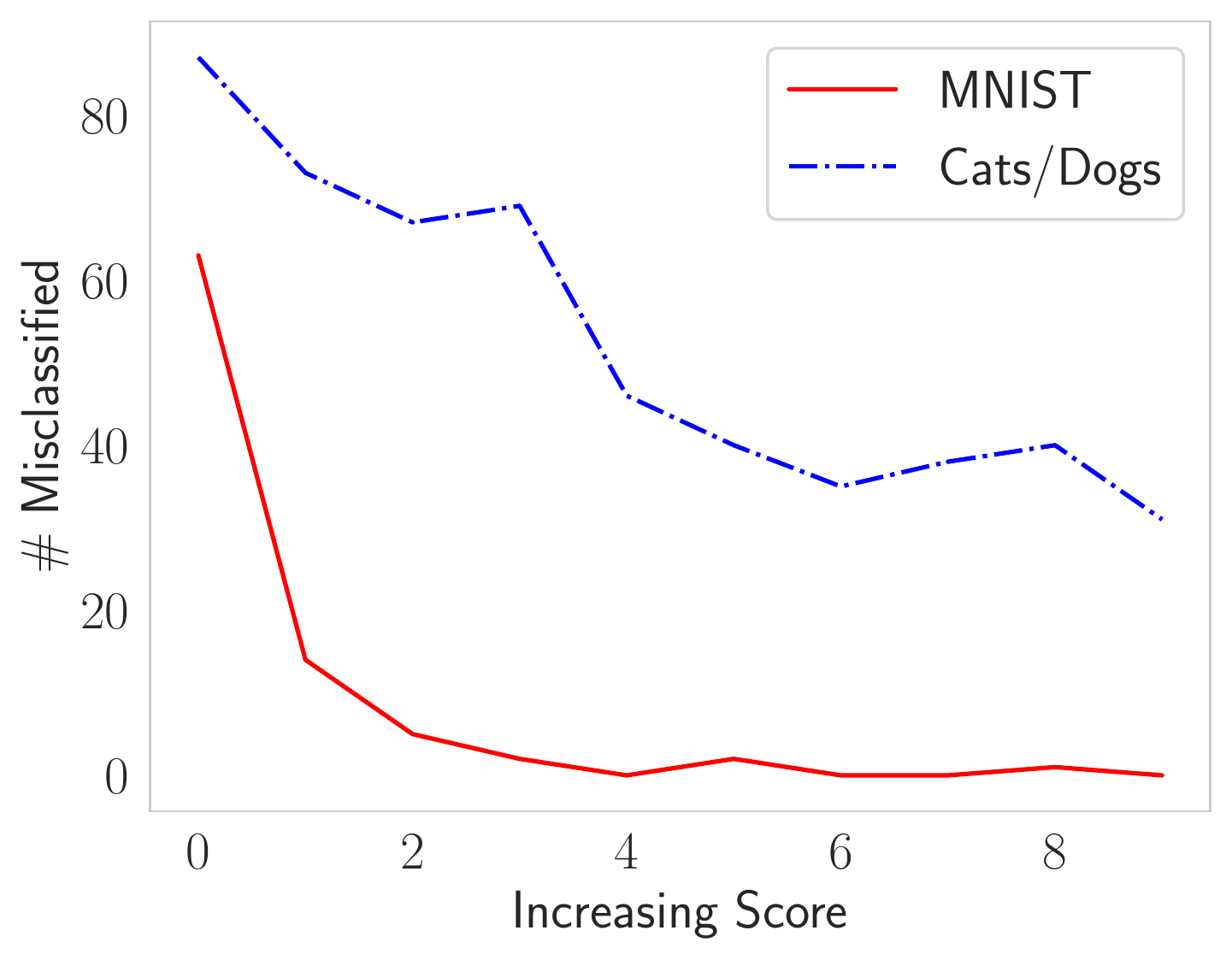}
		\caption{Misclassification by score\label{fig:ScoresC}}
	\end{subfigure}
	\centering\caption{Commonality score and misclassification for the MNIST and the Cats and Dogs test sets\label{fig:Scores}}
\end{figure*}

To identify a correlation between commonality score and the misclassification rate, we divided the testing set into ten groups of equal size maintaining the commonality score ordering. In this way, group 0 had the lowest scoring samples and group 9 the highest.
For each group we then plotted the number of misclassifications in the group. The resulting plot, presented in Figure~\ref{fig:ScoresC},  shows a clear reduction in the misclassification rate as the commonality score increases.

When we consider all of the samples in the MNIST test set we obtain an
accuracy of $0.9913$ and a misclassification rate of $(1-0.9913) = 0.0087$.
However, from $356$ samples which are identified as outliers with a low commonality score in Figure~\ref{fig:ScoresB}, $40$ are misclassified, giving a misclassification rate of $0.1123$. This misclassification rate is almost $13$ times higher than that expected for the model. Repeating this procedure for the Cats/Dogs data, we note that while the misclassification rate for all samples is $0.1054$, the rate associated with the 190 outliers identified is almost double, i.e., $0.2053$.

From these results we can see that a low commonality score is a strong indicator that samples are more likely to be misclassified for the models considered. In the next section we will show how manual inspection of those samples confirms that rare subclasses are indeed present and identifiable in these low scoring samples. We also demonstrate how this information can be used to mitigate the impact of rare subclasses.

\subsection{Mitigation of Rare Subclasses during Training}

\begin{figure}[t]
	\centering
	\begin{tabular}{cc}
		%\fboxsep=0mm%padding thickness
%\fboxrule=1pt%border thickness
\setlength\tabcolsep{1.5pt} 
\begin{tabular}{ccccc}
%	\fcolorbox{red}{red}{\includegraphics[scale=0.20]{images/PGF/Rare4s/MNIST_Rare4s-img0.png} }&
	\includegraphics[scale=0.20]{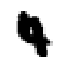}&
	\includegraphics[scale=0.20]{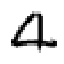} &
	\includegraphics[scale=0.20]{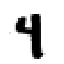} &
	\includegraphics[scale=0.20]{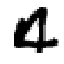} &
	\includegraphics[scale=0.20]{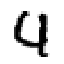} \\
	\includegraphics[scale=0.20]{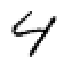} &
	\includegraphics[scale=0.20]{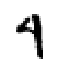} &
	\includegraphics[scale=0.20]{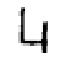} &
	\includegraphics[scale=0.20]{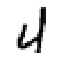} &
	\includegraphics[scale=0.20]{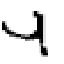} \\
	\includegraphics[scale=0.20]{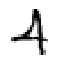} &
	\includegraphics[scale=0.20]{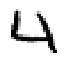} &
	\includegraphics[scale=0.20]{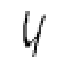} &
	\includegraphics[scale=0.20]{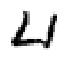} &
	\includegraphics[scale=0.20]{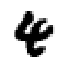} \\
	\includegraphics[scale=0.20]{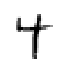} &
	\includegraphics[scale=0.20]{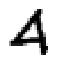} &
	\includegraphics[scale=0.20]{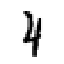} &
	\includegraphics[scale=0.20]{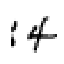} &
	\includegraphics[scale=0.20]{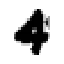} \\
	\includegraphics[scale=0.20]{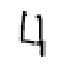} &
	\includegraphics[scale=0.20]{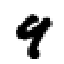} &
	\includegraphics[scale=0.20]{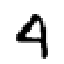} &
	\includegraphics[scale=0.20]{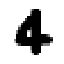} &
	\includegraphics[scale=0.20]{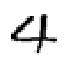} 
\end{tabular}	
 &
		\setlength\tabcolsep{1.5pt} 
\begin{tabular}{ccccc}
	\includegraphics[scale=0.20]{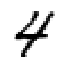} &
	\includegraphics[scale=0.20]{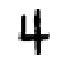} &
	\includegraphics[scale=0.20]{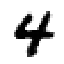} &
	\includegraphics[scale=0.20]{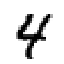} &
	\includegraphics[scale=0.20]{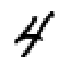} \\
	\includegraphics[scale=0.20]{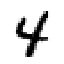} &
	\includegraphics[scale=0.20]{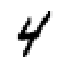} &
	\includegraphics[scale=0.20]{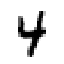} &
	\includegraphics[scale=0.20]{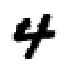} &
	\includegraphics[scale=0.20]{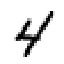} \\
	\includegraphics[scale=0.20]{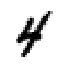} &
	\includegraphics[scale=0.20]{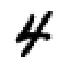} &
	\includegraphics[scale=0.20]{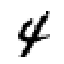} &
	\includegraphics[scale=0.20]{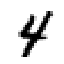} &
	\includegraphics[scale=0.20]{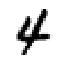} \\
	\includegraphics[scale=0.20]{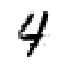} &
	\includegraphics[scale=0.20]{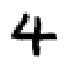} &
	\includegraphics[scale=0.20]{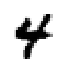} &
	\includegraphics[scale=0.20]{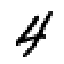} &
	\includegraphics[scale=0.20]{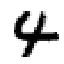} \\
	\includegraphics[scale=0.20]{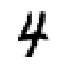} &
	\includegraphics[scale=0.20]{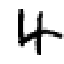} &
	\includegraphics[scale=0.20]{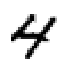} &
	\includegraphics[scale=0.20]{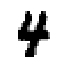} &
	\includegraphics[scale=0.20]{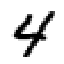} 
\end{tabular}	
\\
		a) Least common fours &
		b) Most common fours \\
		\setlength\tabcolsep{1.5pt} 
\begin{tabular}{ccccc}
	\includegraphics[scale=0.20]{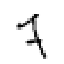} &
	\includegraphics[scale=0.20]{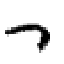} &
	\includegraphics[scale=0.20]{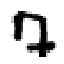} &
	\includegraphics[scale=0.20]{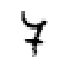} &
	\includegraphics[scale=0.20]{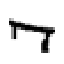} \\
	\includegraphics[scale=0.20]{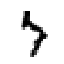} &
	\includegraphics[scale=0.20]{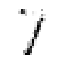} &
	\includegraphics[scale=0.20]{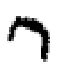} &
	\includegraphics[scale=0.20]{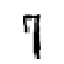} &
	\includegraphics[scale=0.20]{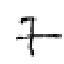} \\
	\includegraphics[scale=0.20]{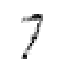} &
	\includegraphics[scale=0.20]{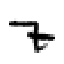} &
	\includegraphics[scale=0.20]{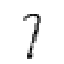} &
	\includegraphics[scale=0.20]{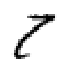} &
	\includegraphics[scale=0.20]{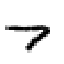} \\
	\includegraphics[scale=0.20]{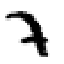} &
	\includegraphics[scale=0.20]{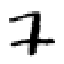} &
	\includegraphics[scale=0.20]{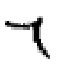} &
	\includegraphics[scale=0.20]{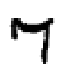} &
	\includegraphics[scale=0.20]{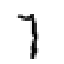} \\
	\includegraphics[scale=0.20]{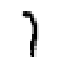} &
	\includegraphics[scale=0.20]{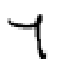} &
	\includegraphics[scale=0.20]{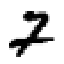} &
	\includegraphics[scale=0.20]{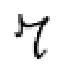} &
	\includegraphics[scale=0.20]{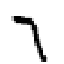} 
\end{tabular}	
 &
		\setlength\tabcolsep{1.5pt} 
\begin{tabular}{ccccc}
	\includegraphics[scale=0.20]{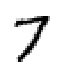} &
	\includegraphics[scale=0.20]{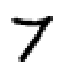} &
	\includegraphics[scale=0.20]{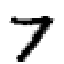} &
	\includegraphics[scale=0.20]{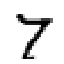} &
	\includegraphics[scale=0.20]{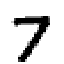} \\
	\includegraphics[scale=0.20]{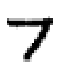} &
	\includegraphics[scale=0.20]{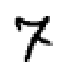} &
	\includegraphics[scale=0.20]{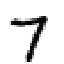} &
	\includegraphics[scale=0.20]{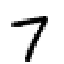} &
	\includegraphics[scale=0.20]{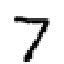} \\
	\includegraphics[scale=0.20]{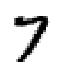} &
	\includegraphics[scale=0.20]{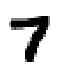} &
	\includegraphics[scale=0.20]{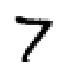} &
	\includegraphics[scale=0.20]{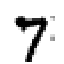} &
	\includegraphics[scale=0.20]{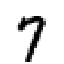} \\
	\includegraphics[scale=0.20]{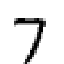} &
	\includegraphics[scale=0.20]{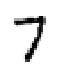} &
	\includegraphics[scale=0.20]{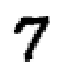} &
	\includegraphics[scale=0.20]{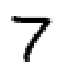} &
	\includegraphics[scale=0.20]{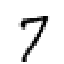} \\
	\includegraphics[scale=0.20]{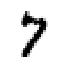} &
	\includegraphics[scale=0.20]{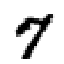} &
	\includegraphics[scale=0.20]{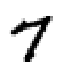} &
	\includegraphics[scale=0.20]{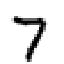} &
	\includegraphics[scale=0.20]{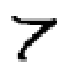} 
\end{tabular}	
\\
		c) Least common sevens &
		d) Most common sevens
	\end{tabular}	
	\caption{The 25 samples with the lowest/highest commonality scores from the MNIST testing data set for digits `4' and `7'\label{fig:digits}}
	\vspace{-5mm}
\end{figure}

\begin{figure*}
\centering
\begin{subfigure}{0.32\textwidth}
\centering
		\setlength\tabcolsep{1.5pt} 
\begin{tabular}{ccccc}
	\includegraphics[scale=0.40]{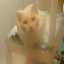} &
	\includegraphics[scale=0.40]{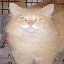} &
	\includegraphics[scale=0.40]{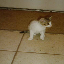} &
	\includegraphics[scale=0.40]{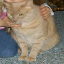} &
	\includegraphics[scale=0.40]{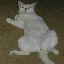} \\
	\includegraphics[scale=0.40]{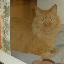} &
	\includegraphics[scale=0.40]{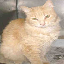} &
	\includegraphics[scale=0.40]{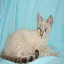} &
	\includegraphics[scale=0.40]{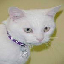} &
	\includegraphics[scale=0.40]{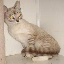} \\
	\includegraphics[scale=0.40]{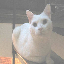} &
	\includegraphics[scale=0.40]{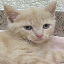} &
	\includegraphics[scale=0.40]{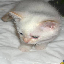} &
	\includegraphics[scale=0.40]{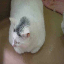} &
	\includegraphics[scale=0.40]{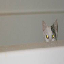} \\
	\includegraphics[scale=0.40]{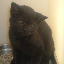} &
	\includegraphics[scale=0.40]{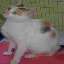} &
	\includegraphics[scale=0.40]{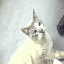} &
	\includegraphics[scale=0.40]{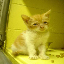} &
	\includegraphics[scale=0.40]{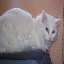} \\
	\includegraphics[scale=0.40]{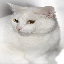} &
	\includegraphics[scale=0.40]{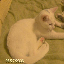} &
	\includegraphics[scale=0.40]{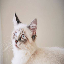} &
	\includegraphics[scale=0.40]{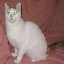} &
	\includegraphics[scale=0.40]{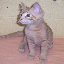} 
\end{tabular}	
 \caption{Lowest scoring cats.\label{fig:25lowestcats}} 		
\end{subfigure}
\begin{subfigure}{0.32\textwidth}
\centering
    \setlength\tabcolsep{1.5pt} 
\begin{tabular}{ccccc}
	\includegraphics[scale=0.40]{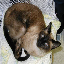} &
	\includegraphics[scale=0.40]{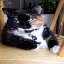} &
	\includegraphics[scale=0.40]{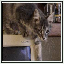} &
	\includegraphics[scale=0.40]{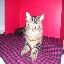} &
	\includegraphics[scale=0.40]{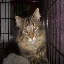} \\
	\includegraphics[scale=0.40]{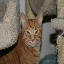} &
	\includegraphics[scale=0.40]{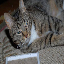} &
	\includegraphics[scale=0.40]{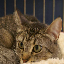} &
	\includegraphics[scale=0.40]{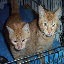} &
	\includegraphics[scale=0.40]{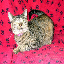} \\
	\includegraphics[scale=0.40]{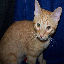} &
	\includegraphics[scale=0.40]{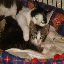} &
	\includegraphics[scale=0.40]{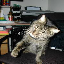} &
	\includegraphics[scale=0.40]{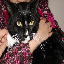} &
	\includegraphics[scale=0.40]{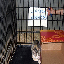} \\
	\includegraphics[scale=0.40]{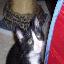} &
	\includegraphics[scale=0.40]{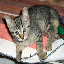} &
	\includegraphics[scale=0.40]{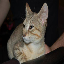} &
	\includegraphics[scale=0.40]{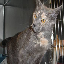} &
	\includegraphics[scale=0.40]{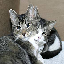} \\
	\includegraphics[scale=0.40]{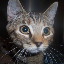} &
	\includegraphics[scale=0.40]{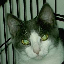} &
	\includegraphics[scale=0.40]{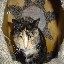} &
	\includegraphics[scale=0.40]{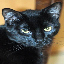} &
	\includegraphics[scale=0.40]{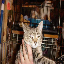} 
\end{tabular}	

 \caption{Highest scoring cats.\label{fig:25highestcats}}
\end{subfigure}
\begin{subfigure}{0.32\textwidth}
\centering
		\setlength\tabcolsep{1.5pt} 
\begin{tabular}{ccccc}
	\includegraphics[scale=0.40]{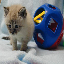} &
	\includegraphics[scale=0.40]{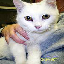} &
	\includegraphics[scale=0.40]{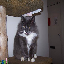} &
	\includegraphics[scale=0.40]{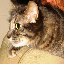} &
	\includegraphics[scale=0.40]{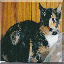} \\
	\includegraphics[scale=0.40]{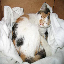} &
	\includegraphics[scale=0.40]{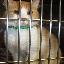} &
	\includegraphics[scale=0.40]{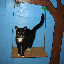} &
	\includegraphics[scale=0.40]{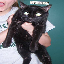} &
	\includegraphics[scale=0.40]{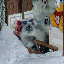} \\
	\includegraphics[scale=0.40]{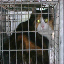} &
	\includegraphics[scale=0.40]{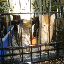} &
	\includegraphics[scale=0.40]{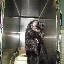} &
	\includegraphics[scale=0.40]{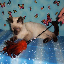} &
	\includegraphics[scale=0.40]{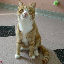} \\
	\includegraphics[scale=0.40]{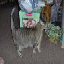} &
	\includegraphics[scale=0.40]{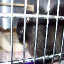} &
	\includegraphics[scale=0.40]{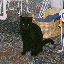} &
	\includegraphics[scale=0.40]{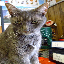} &
	\includegraphics[scale=0.40]{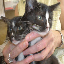} \\
	\includegraphics[scale=0.40]{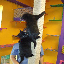} &
	\includegraphics[scale=0.40]{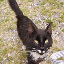} &
	\includegraphics[scale=0.40]{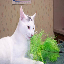} &
	\includegraphics[scale=0.40]{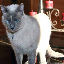} &
	\includegraphics[scale=0.40]{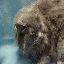} 
\end{tabular}	

	\caption{Randomly sampled misclassified cats.\label{fig:MCOnlyCats}}
\end{subfigure}
\caption{Using the commonality score to identify rare subclasses in the cats/dogs data set.}
\end{figure*}

\noindent
\textbf{Rare subclass detection. }
Having identified the samples with a low commonality score, we next examined whether they could be used to identify rare subclasses. To that end, 
we selected the 25 samples with the lowest and highest commonality scores. The sets of these samples for digits `4' and `7' from the MNIST data set are shown in Figure~\ref{fig:digits}, and those for cats in the Cats/Dogs data set are shown in  Figures~\ref{fig:25lowestcats} and~\ref{fig:25highestcats}.

To identify actionable features associated with rare subclasses, we showed these image sets to nine groups of two or three volunteers each (20~people in total).\footnote{The volunteers included support staff, PhD students, postdoctoral researchers and academics from our Computer Science Department.} The volunteers were asked to identify features present in the 25-sample set with the lowest commonality scores and absent from the 25-sample set with the highest commonality scores. The participants in this study were informed that the first set contained images underrepresented in the DNN training set, and that their responses would aid us in gathering more images of this type.~%A table containing the user responses is available on our supporting website~\url{ourwebsite.com}.
Their responses are reported in Table~\ref{tab:Responses}.

\begin{table*}[t]
	\centering
	\caption{User responses when shown the images with low and high commonality scores (Figures~\ref{fig:digits},~\ref{fig:25lowestcats} and~\ref{fig:25highestcats}).\label{tab:Responses}}
	\footnotesize
	\sffamily
	\begin{tabular}{cp{5.0cm}p{5.6cm}p{5.2cm}} \toprule
		ID & 4s & 7s & Cats \\ \midrule
		1 &  Closed style (rather than open for more common).& Left slope. Emphasised/strong horizontal line. Pronounced top vertical bit & \\\midrule
		2 & Drawn with a triangle & Include horizontal lines. "Serif" font& White fur. Small\\\midrule
		3 & Untidy. Closed Triangle. Extra squiggles. (More common slope right) &Horizontal bar. Serifs. Less Angular & Light coloured. Less contrast to background/cat. Parts of cats\\\midrule
		4 &Some look like 'A's. Extra edge beyond the number. & Some are not 7s. Short top.& Stretched photos. Poor quality photos.\\\midrule
		5 &4 with a triangle form. Stroke does not cut through the stem midway, too high or too low. & Horizontal stroke in the stem of the number. Numbers are curved, not sharp angles. Tail at the base. Stroke at the start.& White cats\\\midrule
		6 & Triangle fours. Small tails &Curved upper horizontal line. Horizontal middle line. Tail flick. Negative gradient slope. & Almost all light coloured. Strange positions.\\\midrule
		7 & Different format of four.& Some incorrectly labelled. Irregular shape.& Not clear with respect to the background\\\midrule
		8 & Using triangle form. Not  sloping right. 'wrong' proportions. Some are excessively thick.&Extra detail. Extra long or short strokes. Top stroke not flat. & White cats/light colours. Light backgrounds. Off angle. Not looking straight at the camera. Strange poses.\\\midrule
		9 & Thickness. Missing the stalk.& Exaggerated curves. The cross through the seven. Slant to the left. Slightly broken lines.&Lighter colours - both the cats and the backgrounds. \\ \bottomrule
	\end{tabular}
	\vspace{-5mm}
\end{table*}

For digit `4', eight of the nine respondent groups noted that the less common fours used the triangle form, whilst it was more common to have an open four. For digit `7', six of the groups deemed that the horizontal bar was present more often in the digits with a low commonality score. For the images of cats shown, six of the nine groups noted that the less common cats were light in colour. 

To confirm that the features identified by our respondents were indeed rare, we used two approaches. For the MNIST data set, we implemented a simple software tool that presented each of the digits labelled `4' and `7' in the training set to a human checker.  We then used the tool to identify all digit `4' samples of triangular form, and all digit `7' samples with a horizontal bar. The digits identified were then counted, and we noted that only 377 of the 5842 fours in the training set were of triangular form (6\%); and only 802 out of the 6265 sevens included a horizontal bar (13\%). Thus, our experiment correctly determined rare MNIST subclasses of `4' and '7'.

For the cats and dogs data set, a different validation approach was taken: we randomly selected sets of 25 images from the training images labelled `cat', and counted the lightly coloured cat images in each set. We repeated this procedure 30 times, and found that each sample of 25 randomly selected cat images contained, on average, only 2.3 images of lightly coloured cats.
On this basis, we estimate that lightly coloured cats represent under 10\% of the cat samples, and therefore are a rare subclass of `cat' in the Cats and Dogs data set.

\vspace*{2mm}\noindent
\textbf{Comparison to detection using missclassified samples.} A natural question is whether rare subclasses cannot also be identified using the sets of misclassified samples obtained from the traditional testing of DNN models. 
To answer this question, we manually compared a set of correctly classified samples and a set of misclassified images. 
We randomly selected 25  misclassified cat images (shown in Figure~\ref{fig:MCOnlyCats}) from the Cats and Dogs test data set. When compared to the low commonality score images from Figure~\ref{fig:25lowestcats}, we note that the light-coloured cats no longer dominate the misclassified subset; indeed, it is difficult to identify distinguishing features for the cats which are misclassified. This analysis suggests that the commonality score is indeed necessary for the identification of rare subclasses for this data set.

\if 0
\begin{figure*}
	\centering
	\begin{tabular}{cc}
		\input{sections/MCCatsTable.tex}&\input{sections/CorrectCatsTable.tex}\\
		25 misclassified cats from the testing set &
		25 correctly classified cats from the testing set
	\end{tabular}	
	\caption{Images randomly sampled from the set of misclassified and correctly classified cats in order to identify features which lead to misclassification without using the commonality score.\label{fig:MCOnlyCats}}
\end{figure*}
\fi

\vspace*{2mm}\noindent
\textbf{Training-time mitigation.} Having identified a set of features associated with rare subclasses, we could undertake action to compensate for these missing features. Two  approaches can be used to balance the number of `rare' samples in the training data. These approaches are described next. 

\begin{figure}
	\begin{subfigure}[b]{0.49\linewidth}
		\centering
		\includegraphics[width=0.6\linewidth]{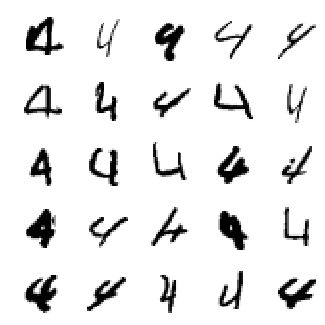}
		\caption{Least common 4s\label{fig:least_4_1K}}
	\end{subfigure}
	\begin{subfigure}[b]{0.49\linewidth}
		\centering
		\includegraphics[width=0.6\linewidth]{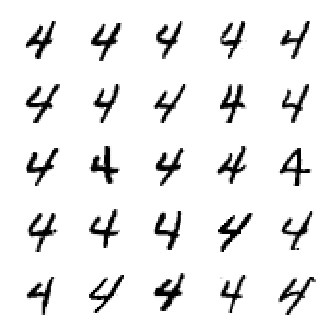}
		\caption{Most common 4s\label{fig:most_4_1k}}
	\end{subfigure}
	\begin{subfigure}[b]{0.49\linewidth}
		\centering
		\includegraphics[width=0.6\linewidth]{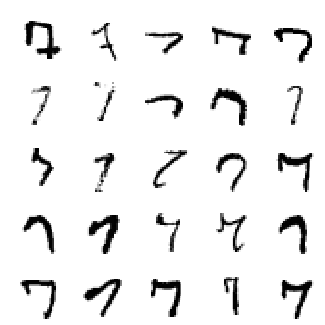}
		\caption{Least common 7s\label{fig:least_7_1K}}
	\end{subfigure}
	\begin{subfigure}[b]{0.49\linewidth}
		\centering
		\includegraphics[width=0.6\linewidth]{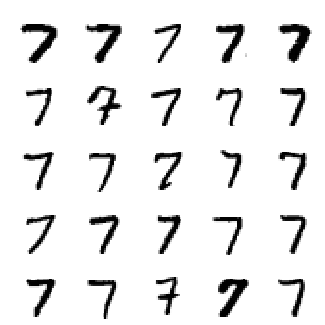}
		\caption{Most common 7s\label{fig:most_7_1k}}
	\end{subfigure}
	\caption{The 25 samples with the lowest/highest commonality score from the MNIST testing set with oversampling\label{fig:Oversampled}}
	\vspace{-4mm}
\end{figure}

For the MNIST data set, we used a manual inspection process to identify samples in the training set that possessed the features identified as rare. A new data set was then constructed in which additional instances of these rare samples were added by oversampling.
First we added an additional 1,000 samples of the triangular fours before creating a new model. The test images of digit `4' with the lowest and highest commonality scores for the new model are shown in~Figure~\ref{fig:least_4_1K} and~\ref{fig:most_4_1k}. When compared to Figure~\ref{fig:digits}, the number of closed form fours has reduced from 8 to 4, and a triangular `4' has now appeared in the common set. We also note that the uniformity of the digits and 'cleanliness' of the handwriting remain distinguishing features of high commonality score samples. 

For the 7s, the addition of 1000 samples with a horizontal bar to the training set did not have this effect. This could be due to the fact that the 7 is a simpler geometric shape than the 4s. We did, however, obtain this effect as more samples were added. Figure~\ref{fig:Oversampled} shows the effect of adding 4,000 rare `7' samples: the horizontal bar from the `7' appears less in the lowest commonality score samples, and appears twice in the most common samples. Moreover, with 4,000 samples added, roughly 48\% of the training samples having a horizontal bar. 

\begin{figure}
	\centering
	\setlength\tabcolsep{1.5pt} 
\begin{tabular}{ccccc}
	\includegraphics[scale=0.70]{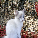} &
	\includegraphics[scale=0.70]{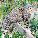} &
	\includegraphics[scale=0.70]{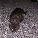} &
	\includegraphics[scale=0.70]{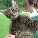} &
	\includegraphics[scale=0.70]{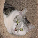} \\
	\includegraphics[scale=0.70]{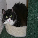} &
	\includegraphics[scale=0.70]{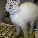} &
	\includegraphics[scale=0.70]{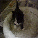} &
	\includegraphics[scale=0.70]{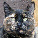} &
	\includegraphics[scale=0.70]{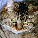} \\
	\includegraphics[scale=0.70]{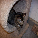} &
	\includegraphics[scale=0.70]{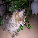} &
	\includegraphics[scale=0.70]{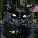} &
	\includegraphics[scale=0.70]{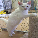} &
	\includegraphics[scale=0.70]{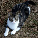} \\
	\includegraphics[scale=0.70]{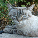} &
	\includegraphics[scale=0.70]{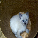} &
	\includegraphics[scale=0.70]{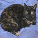} &
	\includegraphics[scale=0.70]{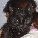} &
	\includegraphics[scale=0.70]{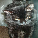} \\
	\includegraphics[scale=0.70]{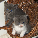} &
	\includegraphics[scale=0.70]{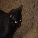} &
	\includegraphics[scale=0.70]{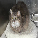} &
	\includegraphics[scale=0.70]{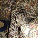} &
	\includegraphics[scale=0.70]{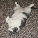} 
\end{tabular}	

	\caption{The 25 lowest scoring cats after the addition of additional training samples for the rare subclass ``white cats"\label{fig:postCats}}
	\vspace{-8mm}
\end{figure}

For the Cats and Dogs data set, having identified light-coloured cats as rare, we used an automated image search to retrieve 1000 images with the keywords ``white cat" from the Internet. We then manually examined the returned set to removed spurious results. This gave us 705 images of white and light-coloured cats, which we labelled as cats and added to the original training data. Having augmented our training set, we then  trained a new model with the structure and training hyper-parameters unchanged.

 Figure~\ref{fig:postCats} shows the 25 cats with the lowest commonality score from the test set for this new model. We note that the images are no longer predominantly light in colour. Many of these new samples are patterned and feature cats on a brown rather than light coloured backgrounds. 
 
\begin{figure*}[t]
	\begin{subfigure}[b]{0.30\linewidth}
		\centering
		\includegraphics[width=\linewidth]{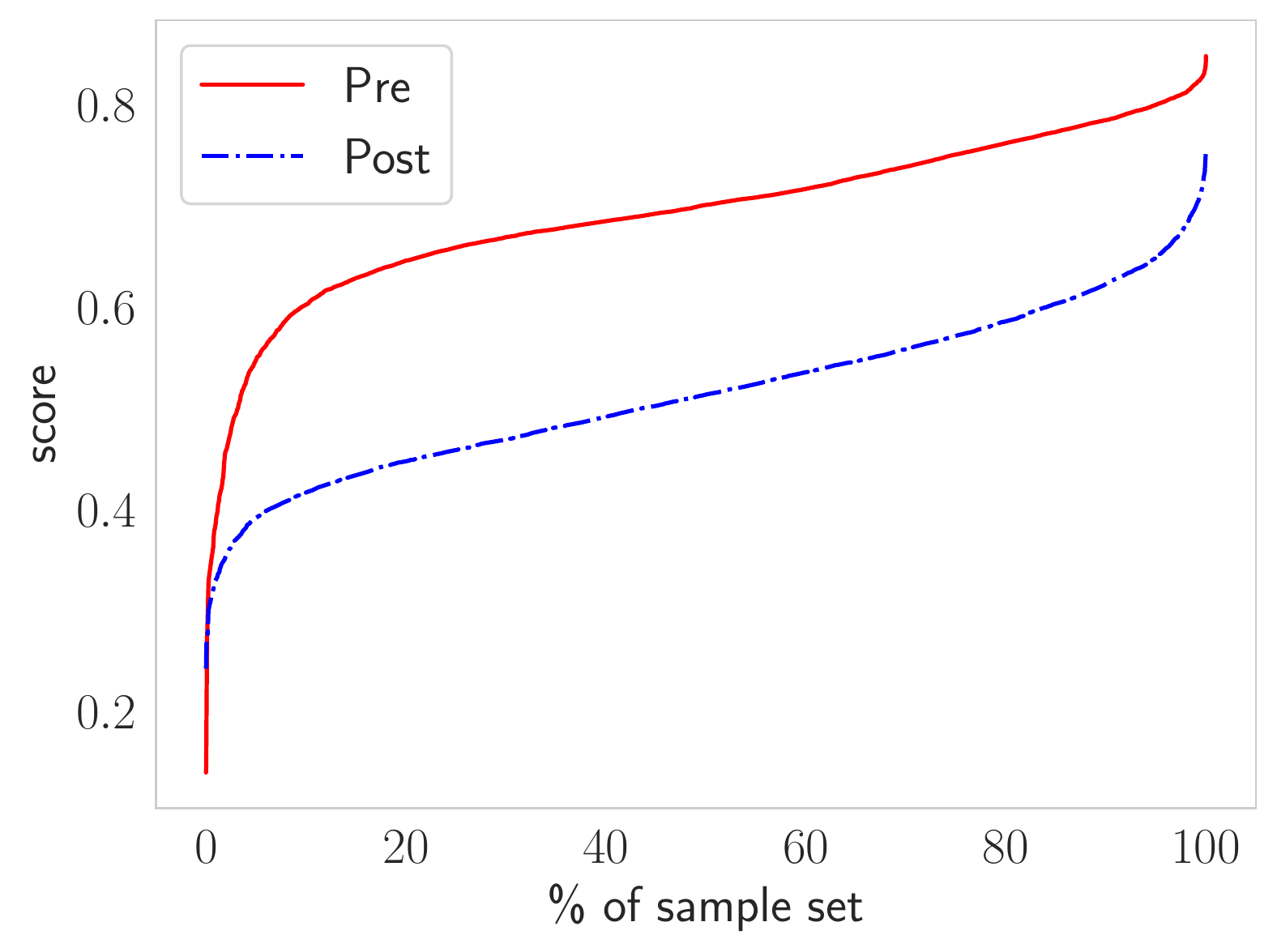}
		\caption{Score of samples in ascending order\label{fig:Scores_Post_A}}
	\end{subfigure}
	\begin{subfigure}[b]{0.30\linewidth}
		\centering
		\includegraphics[width=\linewidth]{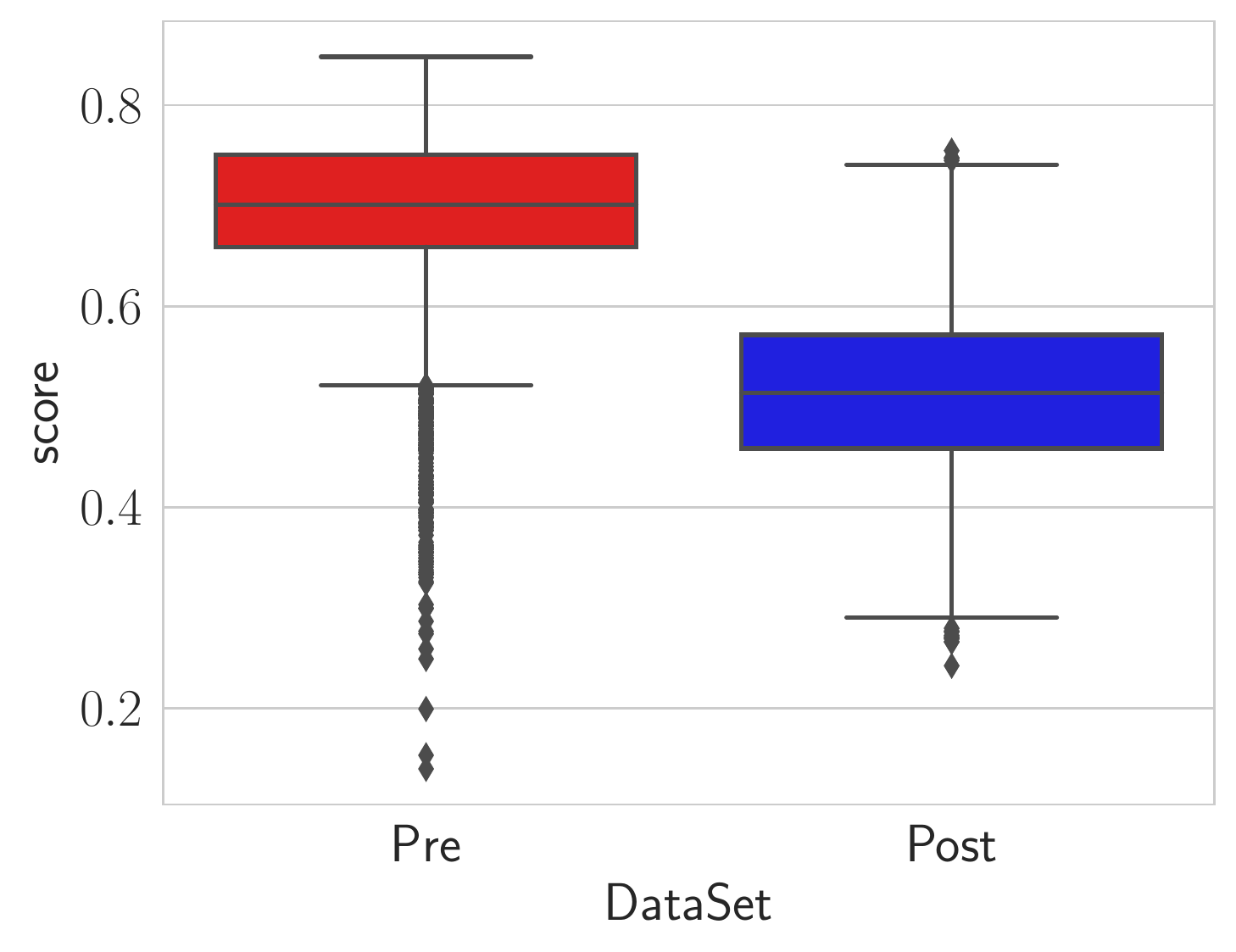}
		\caption{Commonality score with outliers\label{fig:Scores_Post_B}}
	\end{subfigure}
	\begin{subfigure}[b]{0.30\linewidth}
		\centering
		\includegraphics[width=\linewidth]{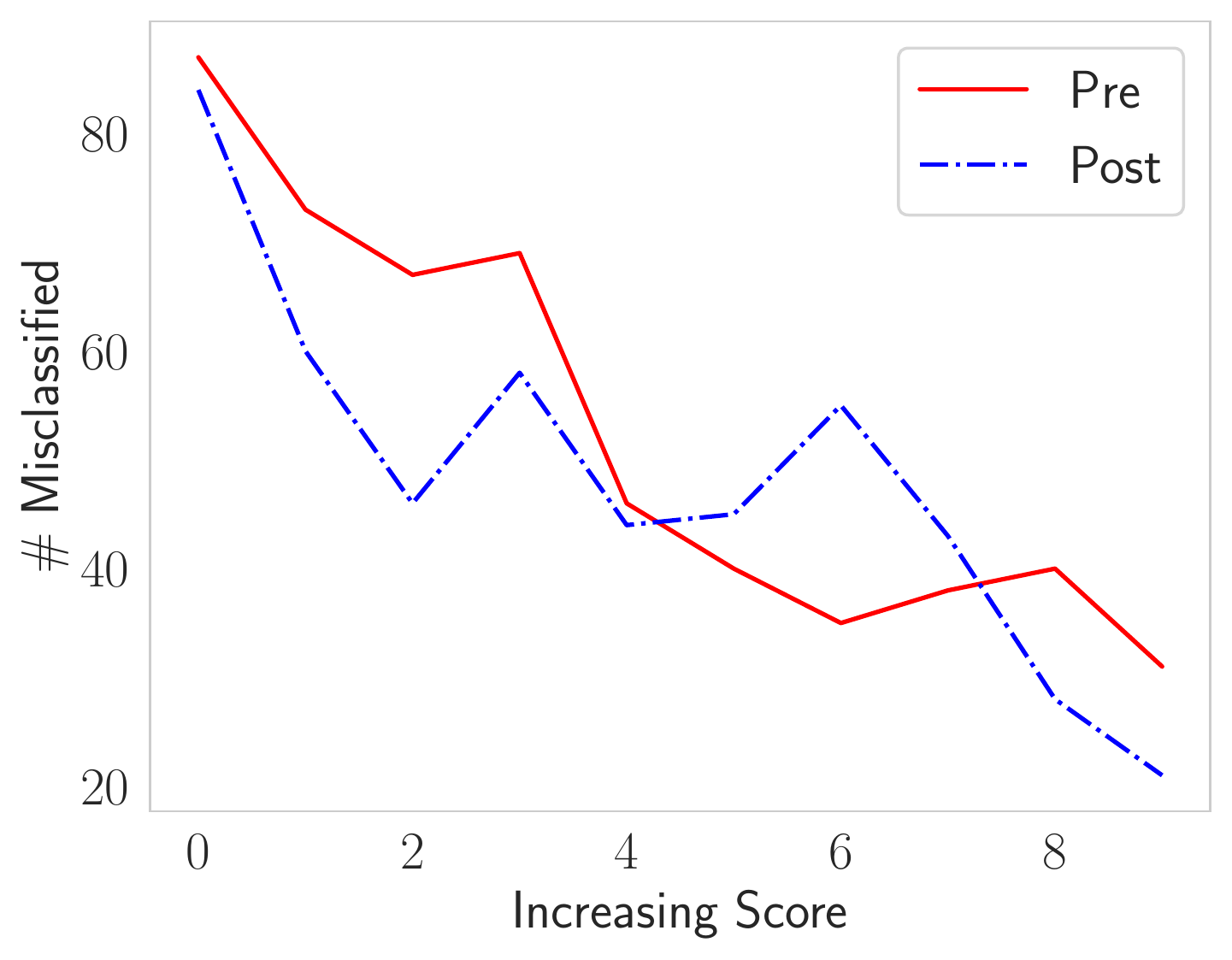}
		\caption{Misclassification by score\label{fig:Scores_Post_C}}
	\end{subfigure}
	\centering\caption{Commonality score and misclassification for the Cats and Dogs test set pre and post training data augmentation\label{fig:CatsDogs_Scores_Post}}
	\vspace{-5mm}
\end{figure*}

\vspace{2mm}Figure~\ref{fig:CatsDogs_Scores_Post} compares the commonality scores for our new Cats and Dogs model to those for the model without an augmented training set. We note that the commonality score has been compressed with the difference between the highest to lowest scores reduced from 0.71 to 0.51.  The box plot shows a reduction in the number of outliers, from 190 to 10 samples. We can also see that the misclassification rate still shows a general tendency to reduce as the commonality score increases.

Finally, we examined the effect that adding additional samples to the training set had on those samples previously identified as outliers. Of the 190 outliers identified for the original model, 144 were cats and 46 were dogs. From these outliers, 19 cats and 14 dogs were misclassified.
After additional training samples were added, all 14 of the dogs continued to be misclassified. In contrast, for the cats, nine of the original 19 misclassified outliers were correctly identified by the new model (and two cats that were previously classified correctly became misclassified). Table~\ref{tab:postModel} shows the accuracy of the model with respect to the test data before and after the addition of ``white cats''. We can see that overall the accuracy of the model increases, with the number of samples misclassified reducing by approximately 8\%. Although it may seem surprising that the reduction in misclassification rate is more pronounced for the class 'dog', this is not unexpected for a binary classifier since improving the specificity of one class necessarily improves the accuracy of both classes. 

\begin{table}[t]
	\centering
\caption{Model comparison pre and post ``white cats''\label{tab:postModel}}
	\begin{tabular}{ccc} \toprule
		& Original Model & Updated Model \\ \midrule
		Test Accuracy & 0.8946 &  0.9030 \\
		\# misclassified & 526 & 484 \\
		\# cats & 224 & 222 \\
		\# dogs & 302 & 262 \\ \bottomrule
	\end{tabular}
\vspace{-5mm}
\end{table}

We have shown that by using the commonality metric to inform the collection of additional data we were able to reduce the misclassification rates for light-coloured cats by the new DNN model, and that our commonality score no longer identifies light coloured images as underrepresented in the training set.

\subsection{Mitigation of rare subclasses at run-time}

Whilst ground truth labels exist at training and testing time, these are typically unavailable at run-time. Even where we are confident that the training set is a ``complete" description of possible inputs at development time, the dynamic nature of evolving open environments means that inputs seen in the future may not be like those seen during training. In these circumstances, the commonality score can be used as an indicator of rarity at run-time, enabling important reductions in  misclassification rates.

To evaluate the use of the commonality score at run-time 
we considered the Cats/Dogs model and 6934 unlabelled test images which had not previously been used. These images were then classified using the original model, i.e. the model before the addition of additional white cats. 

We used the commonality scores of the test samples to compute the DNN trustworthiness threshold~\eqref{eq:outlier} for $k=1.5$. Unsurprisingly, the resulting threshold, $\tau = 0.5151$, corresponds to the knee from the Cats and Dogs commonality score plot in Figure~\ref{fig:ScoresA}.
When applied to the unlabelled test set, 247 images were identified as having a commonality score below the threshold. Each of these images (which we made available at \url{https://www.cs.york.ac.uk/tasp/rarity/}) were then inspected manually, to verify the validity of the  predicted class. 

Since the test accuracy of the model reported by the training and testing procedure is 0.8946 (Table~\ref{tab:postModel}), we would expect to find approximately 26 misclassified samples in a set of 247 images drawn randomly from the test set. Instead, the set of ``untrustworthy'' samples identified by out method contained 42 misclassified samples, i.e., 61\% more samples than expected.  
This coresponds to a model accuracy of only 0.8299 for the 247 samples with commonality scores below the threshold $\tau$.

\section{Related Work\label{sec:Related}}

Considerable work has been undertaken to assess classifiers with respect to fairness and legislative frameworks~\cite{dwork2012fairness,feldman2015certifying,kamiran2012data}. In this work, data features are known a-priori, with protected classes often enshrined in law. 
Using these well-defined class features, it is possible to define a similarity metric against which the performance of a DNN model may be assessed with respect to these known, or partially known, features.
These works consider features that can be mapped to subclass or group features meaningful to the developer, e.g., gender or age. Our work does not require such a clear link to be known in advance. In our cats and dogs classifier the colour of cats was never given as a feature of the training data, and yet a ``discriminated'' group of white cats was identified. As such, our method is able to identify commonly underrepresented classes without a-priori knowledge.

Cheng et al. consider the monitoring of neuron activation patterns in neural networks~\cite{cheng2019runtime} using activation patterns observed during training and a hamming distance to any sample at run-time. This approach requires all observed patterns to be stored in Binary Decision Diagrams, and therefore its scalability is problematic with increases in i) the number of training samples, ii) the size of the network and iii) the hamming distance. The authors state that their approach is limited to patterns with approx.\ 200 neurons, a limitation that does not exist with our approach, where increasing the model size has only a modest impact on the computational cost of our metric. 
The use of a hamming distance also assumes that all neurons in the pattern carry equal importance. In contrast, our metric encodes the frequency with which neurons are activated, and increasing the size of the training set does not increase the complexity of calculation necessary for the commonality metric. We would, however, be interested to explore how patterns of interaction in the neuron activations could be integrated with a metric for commonality.

Anomaly detection is an active research field~\cite{chalapathy2019deep,chandola2009anomaly} and rare subclasses may be regarded as samples that are out-of-distribution (OOD) or outliers with respect to the training distribution.
Supervised approaches to ODD detection ~\cite{hendrycks2016baseline, hendrycks2018deep,  lakshminarayanan2016simple} require outliers to be labelled, which is something that our approach does not require.
In contrast, unsupervised ODD detection learns a density model for the training set, such that any sample with sufficiently low density under the approximated probability distribution is deemed an outlier~\cite{bishop1994novelty}.  Generative deep learning models~\cite{foster2019generative} and variational auto-encoders (VAEs)~\cite{kingma2013auto} have shown some potential in this area, and led to the development of probabilistic auto-encoders~\cite{bohm2020probabilistic} and  hierarchical VAEs~\cite{havtorn2021hierarchical}.
Despite their potential, these approaches involve training additional models for ODD, which needs significant computational resources. Our approach, by contrast, has the advantage to require linear computational time, without the need of training additional models. Moreover, the subclasses identified by our approach, whilst rare, may not be out-of-distribution, and hence may be complementary to OOD methods.
Finally, we showed that our commonality score can be used both at design-time and run-time, which is typically not possible with OOD approaches. 

\section{Conclusions\label{sec:Conclusions}}
Data provide the foundation upon which ML models are built, and without a thorough understanding of the information encoded in these data we are liable to place undue confidence in decisions based on the output of ML components.

When complex neural networks report high levels of accuracy in classification tasks, this may hide performance issues for subclasses of the input set. When this classification underpins important decisions, the systems using such ML models may end up discriminating against these rare subclasses.

We have shown how our approach may be used to support the identification of rare DNN subclasses  for high dimensional input spaces with limited computational effort. Furthermore, we have shown that it is possible to use this information to mitigate possible discrimination by improving the model during training, and to identify higher levels of misclassification than is currently possible at run-time.

In future work we would like to systematically compare our approach to other techniques in this rapidly evolving field of research, and investigate if, by combining techniques, we could improve the process of rarity detection and mitigation further.

\vspace{0.2cm}
{\bfseries\noindent \newline Acknowledgements.} 
This work was funded by the Assuring Autonomy International Programme, and the UKRI project EP/V026747/1 `Trustworthy Autonomous Systems Node in Resilience'.

\bibliographystyle{plain}
\bibliography{rarity}

\end{document}